\documentclass[sigplan,screen]{acmart}
\usepackage{placeins}
\usepackage{graphicx}
\usepackage[export]{adjustbox}
\setcopyright{none}
\renewcommand\footnotetextcopyrightpermission[1]{}

\acmConference[WACV 2024]{{IEEE/CVF} Winter Conference on Applications of Computer Vision}{January 4-8, 2024}
{Waikoloa, HI, USA}

\AtBeginDocument{%
  \providecommand\BibTeX{{%
    \normalfont B\kern-0.5em{\scshape i\kern-0.25em b}\kern-0.8em\TeX}}}

\begin{document}

\title{Camera-Independent Single Image Depth Estimation from Defocus Blur}

\author{Lahiru Wijayasingha}
\email{lnw8px@virginia.edu}
\affiliation{%
  \institution{University of Virginia}
  \country{USA}
}
\author{Homa Alemzadeh}
\email{ha4d@virginia.edu}
\affiliation{%
  \institution{University of Virginia}
  \country{USA}
}
\author{John A. Stankovic}
\email{stankovic@cs.virginia.edu}
\affiliation{%
  \institution{University of Virginia}
  \country{USA}
}


\begin{abstract}
Monocular depth estimation is an important step in many downstream tasks in machine vision. We address the topic of estimating monocular depth from defocus blur which can yield more accurate results than the semantic based depth estimation methods. The existing monocular depth from defocus techniques are sensitive to the particular camera that the images are taken from. We show how several camera-related parameters affect the defocus blur using optical physics equations and how they make the defocus blur depend on these parameters. The simple correction procedure we propose can alleviate this problem which does not require any retraining of the original model. We created a synthetic dataset which can be used to test the camera independent performance of depth from defocus blur models. We evaluate our model on both synthetic and real datasets (DDFF12 and NYU depth V2) obtained with different cameras and show that our methods are significantly more robust to the changes of cameras. Code: \url{https://github.com/sleekEagle/defocus_camind.git}
\end{abstract}

\maketitle

\section{Introduction}

Computer vision based depth estimation has many applications such as augmented and virtual reality (AR and VR) \cite{Ping2019ComparisonSystems}, autonomous robotics \cite{Dong2022TowardsSurvey}, background subtraction, and changing the focus of an image after it was taken \cite{Wang2021BridgingSupervision}\cite{Maximov2020FocusEstimation}\cite{Casser2019DepthVideos}. Techniques such as structure from motion, structure from shading, shape from structured light, shape from defocus blur, depth from focus, multi-view stereo and Time-of-Flight (ToF) sensors can be used to estimate the depth of a scene \cite{Gur2019SingleCues,Watanabe1996Real-timeDefocus}. Active methods such as structured light and ToF sensors need specialized hardware and are power hungry. Stereo techniques measure depth by relying on multiple cameras to take several pictures of the scene. Techniques such as structure-from-motion and depth-from-focus \cite{Yang2022DeepVolume} require several images of a static scene to estimate it's structure. Also, the assumption about static scene does not hold when the scene is changing over time. Furthermore, structure from motion can only recover the depth of a scene up to a scale and cannot measure the absolute depth.

Single image defocus blur based depth estimation is a fairly under-explored topic in the literature \cite{Wang2021BridgingSupervision} which utilizes the phenomena that certain objects in a photo appear more blurred than the others depending on the distance to those objects from the camera. Therefore, measuring the amount of defocus blur at a point of an image can provide a way to recover the depth to the respective point in the real 3D world. As we will show in Section 3, this method is effective for close range depth measurements (typically under 2 to 3 meters). This makes defocus blur-based depth estimation techniques ideal for measuring depth under many situations including in microscopic scenes \cite{Ban2022DepthScenes,Zhang2022ParticleMethod} and measuring depth to hands and nearby objects for a wearable camera.


Single image depth from defocus blur methods are not robust to changes of cameras. As we will show in our experiments, the performance of existing methods degrades significantly when they are trained on images taken from one camera and evaluated on images taken from another camera (even when they both image the same scene). This is due to the fact that different cameras will produce defocus blurs with different characteristics.

In this paper we describe a novel technique to estimate depth from defocus blur in a camera-independent manner. We exploit the optical physics equations that describe the relationships between various camera parameters and the amount of defocus blur. Our method can be used to train a deep learning model in a supervised manner on a dataset containing defocus blurred images taken from a single or multiple camera/s and respective ground truth depth maps. This trained model can be used to predict depth using images taken with a wide range of other cameras with a slight modification to the model (depending on the particular camera parameters of the new camera) and without the need for retraining. We also describe a novel method to estimate the camera parameters of a given camera with an easy to use calibration process. This will be particularly useful when the parameters for a certain camera cannot be obtained (certain manufacturers do not provide all the parameters in the data-sheets and/or the values are only provided as approximations).

Our main \textbf{contributions} are as follows:
\begin{itemize}
    \item We show that depth from defocus technique can measure depth more accurately than the state-of-the-art techniques. 
    \item We show that existing depth from defocus methods are not robust to changes of cameras the images are acquired with.
    \item This paper is the first to device a relationship between defocus blur and the blur created due to pixel binning.
    \item We present a novel depth from defocus blur method which is robust to images taken from a wide range of cameras, given camera parameters that describe a particular camera. 
    \item We present a novel calibration technique to estimate the camera parameters based on several images taken from a given camera.
    \item Our methods have less estimation error than the state-of-the-art when performing depth from blur in a camera-independent manner. The error reduction is around 3cm under the DDFF12 dataset, 7cm under the NYU depth v2 dataset and around 5cm for the synthetic dataset we created. 
\end{itemize}


\section{Related Work}
\subsection{Depth from RGB images}
Estimating depth maps from images can use various characteristics of images such as semantics, stereo matching, blur or differences in blur over a stack of images. \cite{Kong2019DeepMotion,Wang2020360sd-net:Volume,Chang2008PrioritySystem}. Although stereo matching based and blur based depth measurements are seen as completely separate methods, Schechner and Kiryati \cite{Schechner2000DepthThey} showed that both of them can be understood under the same mathematical formulation. A depth map can be estimated for the given image based on the domain knowledge on the structure of the objects in the image embedded in the estimation model \cite{Li2022Omnifusion:Fusion,Mertan2022SingleOverview,Patil2022P3depth:Prior,Bhat2023Zoedepth:Depth,Zhao2023UnleashingPerception}. Methods such as ZoeDepth \cite{Bhat2023Zoedepth:Depth} and VPD \cite{Zhao2023UnleashingPerception} have pushed the state-of-the art to be very accurate in measuring depth. However, a problem with these methods is that the estimated depth is only an approximation based on the structure of the objects. This makes these models sensitive to domain changes \cite{Gur2019SingleCues}. Also, techniques that can recover 3D structure from RGB images such as structure from motion can only estimate relative depths in a given scene \cite{Xian2020Structure-guidedPrediction}. 

\subsection{Depth from defocus blur}

The amount of defocus blur can be used to measure the depth of a scene from images. Since these methods rely more on blur which is a local feature of the image to estimate the depth, they are more robust to domain changes \cite{Gur2019SingleCues}. Shape/depth from focus methods aim to measure depth to a scene given a stack of images of different focus levels. A measure of the sharpness of each pixel of the images over the stack is calculated. The depth of a point is taken as the focus distance with the sharpest pixel. Various methods such as the Laplacian or sum-modified-Laplacian, gray-level variance and gradient magnitude squared were traditionally used to measure the sharpness \cite{Nayar1994ShapeFocus,Subbarao1998SelectingDepth-from-focus}. Modern methods utilize deep learning to automatically learn the sharpness measure from focal stacks \cite{Hazirbas2019DeepFocus,Yang2022DeepVolume,Wang2021BridgingSupervision}. But deep learning based techniques require a large amount of data to train \cite{Maximov2020FocusEstimation}.

Depth from focus methods that use a focal stack of the same scene has several drawbacks. First they assume the scene is static during the time needed to acquire several images with different focus (focal stack). Second, an accurate registration of the images in the focal stack is needed due to focal breathing (slight change of the filed-of-view of the camera due to changes of focal distance) or small movement of the camera and/or the scene \cite{Liu2016DefocusModel}. Therefore, more investigation on depth estimation with a single image is necessary. Depth from defocus/blur rely on measuring the exact blur on a single image to estimate the depth and cannot use the relative variation of sharpness/blurriness of a focal stack. Due to this, depth from blur can be used to estimate the depth from a single blurred image \cite{Anwar2017DepthImage., Subbarao1994DepthApproach, Zhang2021JointConsistency,Gur2019SingleCues, Maximov2020FocusEstimation}. Certain works are also concerned about removing the blur at the same time as estimating depth \cite{Anwar2017DepthImage., Maximov2020FocusEstimation,Pertuz2013GenerationImages}.

Estimating depth from the amount of the blur of a single image is ill-posed. This is due to having two possible depth values for a given blur \cite{Gur2019SingleCues}. Researchers have take two different paths to solve this. One solution is hardware based. One example for this is changing the shape of the aperture (coded aperture) of the camera to a shape that can help avoid the ambiguity. Ikoma et al. \cite{Ikoma2021DepthEstimation} used deep learning to learn the optimal shape for an aperture and came up with a prototype camera to measure depth from blur. Another example is to use a lightfield camera which takes many pictures with closely spaced micro lenses placed inside the camera \cite{Ng2005LightCamera}. The second approach is to use the domain knowledge (e.g. the shape and sizes of objects in the scene) of the scene to remove the ambiguity. Our research falls into this category. Gur and Wolf created a model which can generate the depth map of a scene given the blurred image and the All-in-Focus (AiF) image \cite{Gur2019SingleCues}. They also makes certain assumptions about the shape of the blur circle. Usage of both AiF image and blurred images in making prediction makes this model less useful in certain situations because both of these images are not usually available from regular cameras. Many methods in the literature first estimate the blur of a given blurred image and secondly estimate the depth from the blur. Physical consistency between the estimated blur and depth has been used as a form of domain knowledge by Zhang et al. \cite{Zhang2021JointConsistency}. Lu et al. create two separate models to estimate the blur and the amount of focus (sharpness) of a given blurred image. They claim that this method provides better estimates of depth due to the capability of estimating both blur and the sharpness of an image \cite{Lu2021Self-supervisedClues}. But since sharpness is just the inverse of blur, a question remains that by estimating blur aren't we also estimating the (inverse of) sharpness. Ban et al. \cite{Ban2022DepthScenes} extend depth from blur to microscopic images. Certain works focus just on blur estimation from a blurred image. Tai and Brown use hand crafted features of an image to estimate the blur map \cite{Tai2009SinglePrior} while Zhuo and Sim assume that the edges in the images are step edges \cite{Zhuo2011DefocusImage}. Cun et al. estimated the blur of a given image to separate the blurred and focused areas from a blurred image \cite{Cun2020DefocusDistillation}. While all of the above methods assume that the blur is a single parameter (e.g. Gaussian or disk shape) Liu et al. expand our understanding by introducing a two parameter model. This model is also helpful in removing errors due to pattern edges \cite{Liu2016DefocusModel}. 

\subsection{Camera dependency of depth from blur}
Certain characteristics of blur depend on the camera that is being used to acquire the images. The blurred image of a point has the same shape (but scaled) as the lens aperture. For example, if the aperture is circular, the blur of a point is also circular theoretically. But in practice this is a Gaussian due to diffraction \cite{Subbarao1994DepthApproach}. In this research we assume all the apertures are circular in shape. 

The size of the blur of a point depends on many other parameters of the camera. The f-number, focal length, pixel size of the image sensor (if the camera is digital), camera output scale and focal distance all affect the size of the blur \cite{Gur2019SingleCues} as shown in section 3. Depth from defocus blur techniques estimate the blur of a given image as an intermediate step when estimating the depth. This makes these models sensitive to the variations due to camera parameters. We show evidence supporting this in our evaluation section. But no papers in the literature address this problem. Gur and Wolf \cite{Gur2019SingleCues} use camera parameters in their model to recreate a blurred image. It was not used directly to predict depth and they do not test their model under different cameras. Although Maximov et.al \cite{Maximov2020FocusEstimation} evaluate their model on several simulated datasets generated with different camera parameters they do not explicitly address or propose a solution to this problem.

\section{Approach}
This section starts with a theoretical introduction to estimating depth from defocus blur and establishes the challenges faced by this technique and our solution. 

\subsection{Theory and Techniques}

When imaging a scene with a camera, the points that are not in focus appear blurred and the points that are perfectly in focused appear sharp in the image. This phenomenon is called defocus blurring. To illustrate this, in the left side of the Figure 1, the point $P2$ that is in focus appears as a point in the image plane of the camera. A point $P1$ that is not in focus appears as a blur in the image plane where the pixel intensity is the highest at the center and gradually falls of as we move away. This can be modelled with a 2D Gaussian function as denoted in equation 1 with $\sigma$ as the standard deviation, $x$ and $y$ are image coordinates.

\begin{figure}[t]
  \centering
   \includegraphics[width=1\linewidth]{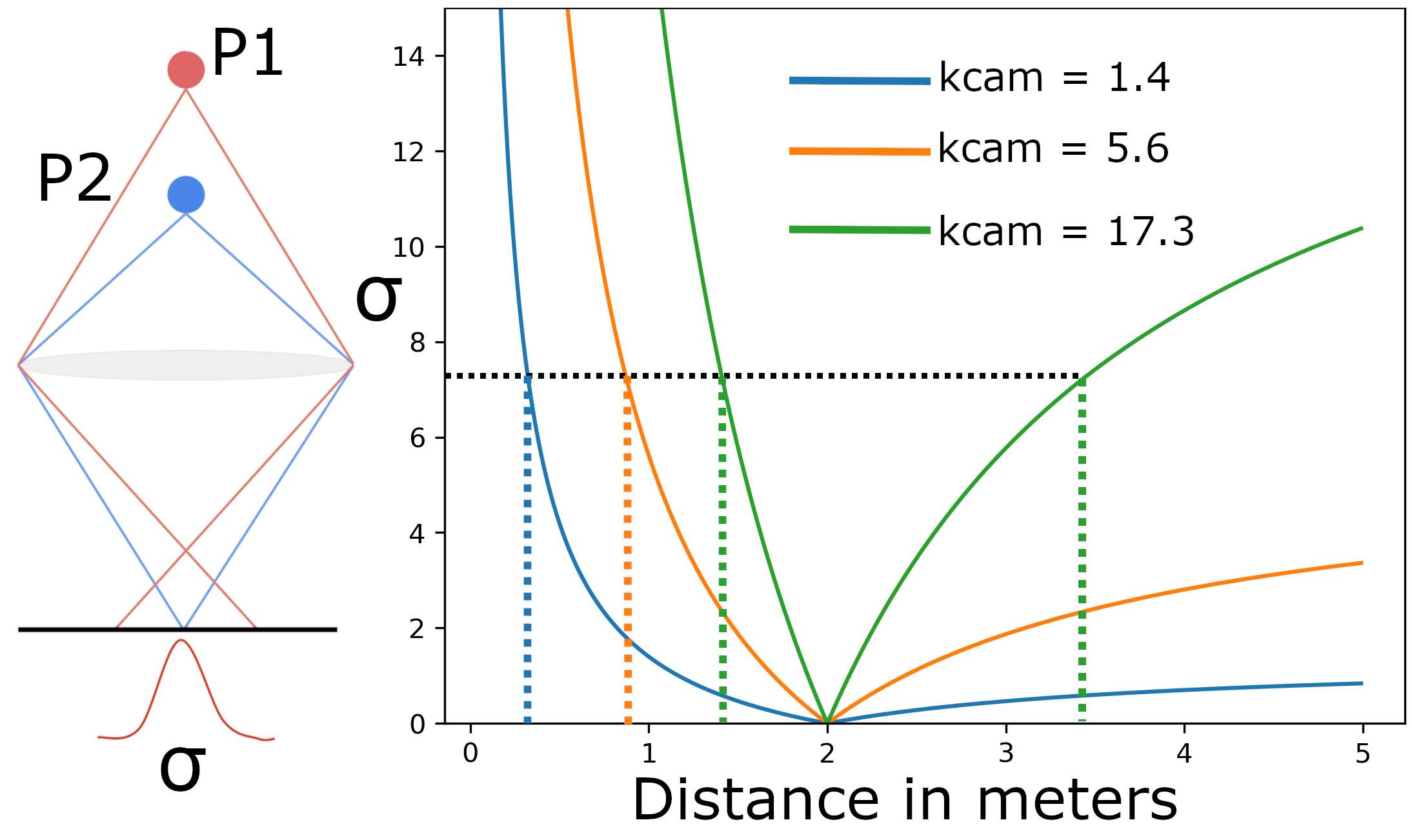}
   \caption{Left:Image formation in a simple camera system. Right:Blur vs. distance}
   \vspace{-0.5cm}
\end{figure}

\begin{equation}
   G(x,y) = \frac{1}{2\pi \sigma} e^{-\frac{1}{2}\frac{x^2+y^2}{\sigma^2}}
  \label{eq:important}
\end{equation}

$\sigma$ depends on the distance to the point $P1$ from the camera center and several other camera dependent factors as shown in equation 2.   

\begin{equation}
   \frac{|s_1-s_2|}{s_2} \cdot \frac{1}{(s_1-f)} \cdot \frac{f^2}{N}\cdot \frac{1}{p} \cdot \frac{out_{pix}}{sensor_{pix}} = k_r \cdot \sigma
  \label{eq:important}
\end{equation}

In equation 2, $s_2$ is the depth (distnace to the )$f$ is the focal length of the camera, $N$ is the f-number, $p$ is the pixel width, $out_{pix}$ is the number of pixels in the final image, $sensor_{pix}$ is the number of pixels in the image sensor, $s_1$ is the focus distance, $k_r$ is a constant that depends on the camera \cite{Subbarao1994DepthApproachb}. Many cameras allow user to change $s_1$ thereby focusing the camera at different distances. we define a camera dependent parameter $k_{cam}$ as shown in equation 3.

\begin{equation}
   \frac{|s_1-s_2|}{s_2} \cdot k_{cam} = \sigma
  \label{eq:important}
\end{equation}

where $k_{cam} = \frac{1}{(s_1-f)} \cdot \frac{f^2}{N}\cdot \frac{1}{p} \cdot \frac{out_{pix}}{sensor_{pix}} \cdot \frac{1}{k_r}$

$G(x,y)$ is the response of the camera system to a point target and is called the point spread function (PSF). We can obtain the defocus blurred image $B(x,y)$ by convolving the the perfectly focused image $F(x,y)$ with PSF $G(x,y)$ as show in equation 4. 

\begin{equation}
   B(x,y) = G(x,y) * F(x,y)
  \label{eq:important}
\end{equation}

Equation 4 explains the blur solely due to defocus blurring. An additional blurring can occur due to various other reasons such as filtering in the camera hardware (e.g. to reduce noise), pixel binning, color filter mosaics, analog/digital image processing, analog to digital conversion, etc. \cite{JeffMeyerandAlexSummersbyImageExplained}. This additional blurring can also be modelled as a convolution with another Gaussian function $Q(x,y)$ having a standard deviation $\gamma$ which we assume to be constant for a given camera stetting. The final image can be obtained by 

\begin{equation}
   I(x,y) = Q(x,y) * G(x,y) * F(x,y)
  \label{eq:important}
\end{equation}
All we can observe is the final image $I$. We show that the combined blurring (from defocus and due to other reasons described above) can also be modelled with a Gaussian PSF (refer the Appendix) and we can estimate the standard deviation of this PSF ($\lambda$) at each pixel in $I$. After estimating $\lambda$ for a given image and when $\gamma$ is known we can obtain $\sigma$ as shown in the equation 6. 

\begin{equation}
   \sigma = \sqrt{\lambda^2 - \gamma^2}
  \label{eq:important}
\end{equation}

Substituting equation 6 into equation 3 we can obtain equation 7.

\begin{equation}
  \frac{|s_1-s_2|}{s_2} \cdot k_{cam} = \sqrt{\lambda^2 - \gamma^2}
  \label{eq:important}
\end{equation}

The right side of the Figure 1 shows the variation of $\sigma$ with different distances ($s_2$) and under different cameras. For a given camera (hence for a given $k_{cam}$), we can estimate the $\sigma$ from a given image and then estimate $s_2$. For certain sections of the curve (e.g. curve of $k_{cam}=17.3$ at the shown value of $\sigma$), estimating $s_2$ is ambiguous since there will be two $s_2$ values for a given $\sigma$. This limitation can be mitigated by using a learning based model to estimate $s_2$. Another observation is that the value of $\sigma$ depends on $k_{cam}$. This poses the main problem that we are addressing in this paper. If a model was trained to estimate depth using data from a camera with one $k_{cam}$, this model will fail to predict the depth accurately for images taken with a camera having a different $k_{cam}$. Furthermore, the sensitivity of $\sigma$ to the distance diminishes as the distance increases. Hence the effectiveness of the defocus blur based depth measurement techniques will also lessen with increasing distance. This limits the effectiveness of depth from defocus blur techniques to close range; as a rule of thumb, to distances less than 2m.

\begin{table}
  \centering
\resizebox{0.5\textwidth}{!}{%
\begin{tabular}{l| c| c c c}
        \hline
        Camera/device & Lens  & f (mm) & N & $K_{cam}$ \\
        \hline
        Cannon EOS Rebel T7 & EF-S              & 18 & 4   & 15.54 \\
                            &                   & 55 & 5.6 & 105.58\\
                            & EF 50mm           & 50 & 1.2 & 406.16 \\
                            & EF 70-300mm       & 70 & 5.6 & 172.35 \\
       \hline
        Nikon D7500         & Nikon AF-S        & 50 & 1.8 & 240.40 \\
                            & AF-S DX NIKKOR    & 18 & 3.5 & 15.76 \\
                            &                   & 18 & 5.6 & 9.85 \\
                            &                   & 55 & 3.5 & 149.98 \\
                            &                   & 55 & 5.6 & 93.73 \\
        \hline
        Sony Alpha 7 IV     & FE PZ 16-35mm     & 16 & 4   & 8.92 \\
                            &                   & 35 & 4   & 43.13 \\
                            &                   & 16 & 22  & 1.62 \\
                            & FE 70-200 mm      & 70 & 2.8 & 250.94 \\
                            &                   & 200& 2.8 & 2196.48 \\
                            &                   & 70 & 22  & 31.93 \\
        \hline
        Google Pixel 7 Pro  & wide              & 25 &1.85 & 50.39 \\
                            & telephoto         & 120&2.55 & 1577.82 \\

        \hline
\end{tabular}}
  \caption{$k_{cam}$ of some popular cameras}
\end{table}
Table 1 shows some camera models and their $k_{cam}$ values based on the particular lens/settings used. Please refer to the Appendix for a more detailed calculation and for $k_{cam}$ values for more cameras/settings.

\subsection{Our solution}

\begin{figure}[t]
  \centering
   \includegraphics[width=1\linewidth]{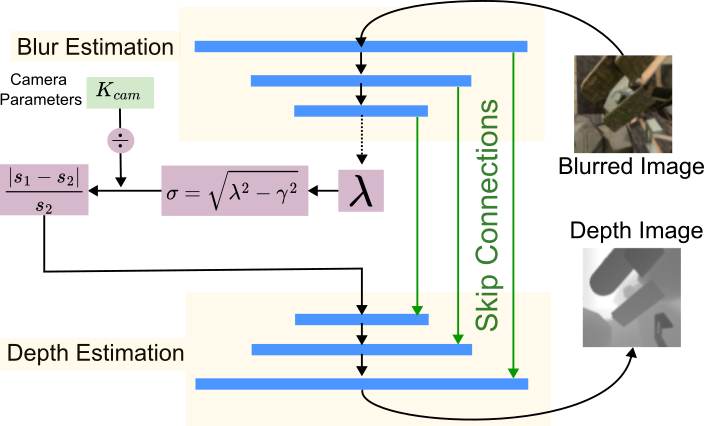}
   \caption{Our Model}
      \vspace{-0.5cm}
\end{figure}

The operation or our model is shown in Figure 2. Both Blur Estimation and Depth estimation sections are CNN based neural networks inspired by the defocusnet \cite{Maximov2020FocusEstimation}. Given a defocus blurred image, the blur estimation model estimates the PSF standard deviation $\lambda$ at each pixel of the image. Then we calculate the  standard deviation of the PSF solely due to defocus blurring $\sigma$ according to equation 6. Next we divide the obtained $\sigma$ with $k_{cam}$ to obtain $\frac{|s_2 - s_1|}{s_2}$ which does not depend on either $k_{cam}$ or $\gamma$. Then $\frac{|s_2 - s_1|}{s_2}$ is sent to the depth estimation model to estimate $s_2$. Note that we require a learning based model such as a CNN to estimate $s_2$ due to the ambiguity explained in the previous section and to impart semantic domain knowledge of the image into the estimation process via the skip connections.

When we train our model, we calculate two types of losses; blur estimation loss and the depth estimation loss. Blur estimation loss ($L_{b}$) is calculated at the prediction of $\lambda$. Ground truth $L_b$ can be obtained with equation 2 with known camera parameters at the training time. Depth prediction loss ($L_{d}$) is calculated comparing the predicted and ground truth depth maps. The final loss is obtained by,
\begin{equation}
  L_{total} = L_d + b\textunderscore weight \cdot L_b
\end{equation}
where $b\textunderscore weight$ is a parameter used to scale $L_b$.

\subsection{Defocus Blur Calibration}
Assume we train our model with images from a certain camera and need to use this already trained model to estimate depth using images from another camera with a different $k_{cam}$ and $\gamma$. In this section we present our novel method that can be used to estimate these parameters for a given camera. We call this method the "Defocus Blur Calibration". Note that defocus blur calibration is different from but requires the conventional camera calibration where camera intrinsics and distortion coefficients are estimated. The steps for defocus blur calibration are as follows.

\begin{enumerate}
\item Fix the focal distance of the camera at $s_1$ (we used $s_1=2m$ in our experiments) and calibrate the camera (in a conventional sense) with a calibration pattern \cite{OpenCV2023Calibration}. We have used an asymmetric circular
pattern as can be seen in Figure 3. Maintain a rough distance of around $\frac{s_1}{2}$ from the camera to the calibration pattern. After this calibration, we can estimate the distance to a given point on the calibration pattern that is visible in a given image. 

\item Capture two images of a circular calibration pattern (preferably the same pattern that was used in step 1) while maintaining a distance of $\frac{s_1}{2}$ (we used $1m$) from the camera to the pattern. The first image is obtained with the camera focused on the pattern ($s_1=1m$) and the second image is obtained while maintaining $s_1=2m$. Since the first image is focused on the calibration pattern, the circles on the pattern will appear sharp as shown in the upper part of Figure 3. The second image will look blurred as shown in the bottom half of the Figure 3. According to Figure 3, the images of circle edges on the focused images have a steeper slope (A Gaussian with a lower std). The slight blurring in these images are solely due to pixel binning. The edges on the blurred images have a more gradual slope. Also the slope becomes even more gradual as $k_{cam}$ is increased.

\item Estimate the std of the Gaussian function of the circle edges from the focused image. We horizontally slice the image of the circle as seen in Figure 3 and obtain the distribution of pixel intensities. These are flat-top Gaussian functions. The flat top nature is due to the intensity being constant inside the circle. It falls gradually at the edges of the circles. We scale these intensity values into the range from zero to one. We consider all the values less than a threshold (we used 0.95) as belonging to the falling edges. We then integrate the resulting distribution (one dimensional Gaussian). According to equation 9, we can estimate $\gamma$ after obtaining the integral $J$ for a constant $y$. 

\begin{equation}
    J = \int_{-\infty}^{\infty} {G(x) dx} = \int_{-\infty}^{\infty} {e^{-\frac{1}{2}\frac{x^2+y^2}{\gamma^2}}} dx  = \gamma \sqrt{2\pi}
  \label{eq:important}
\end{equation}

\item Estimate the std of the Gaussian of the falling edges ($\lambda$) of the circles from defocus blurred images similarly to step 3. Note the the blurring of the defocused images are due to both defocus blurring and pixel binning. We can estimate the std of the Gaussian of the falling edges due to defocus blurring with equation 6. 

\item Estimate the distance to each circle center from camera using the defocus blurred images using the camera intrinsic matrix generated with calibration in step1. This is a well-established procedure that is available in most of the computer vision libraries. 
We can write a separate version of equation 7 for each circle in the calibration pattern. With $\lambda$ and $\gamma$ already estimated, we can estimate the $k_{cam}$ for the given camera using equation 7. Here we have assumed that the distance from the camera to each circle center is approximately equal to the distance to the edges of the circle. This can be justified because the distance to the circles from the camera (around 1m) is much larger than the diameter of the circles (around 4cm in our case). 

\item To improve the accuracy of the estimate, we can repeat steps from 2 to 5 several times. See the evaluation section for further details on the experiments. 

\end{enumerate}

We can estimate the $k_{cam}$ for a given camera with the above steps. The estimated $k_{cam}$ can be used as shown in Figure 2 to predict depth with the images taken from this new camera.  

\begin{figure}
   \includegraphics[width=.8\textwidth,center]{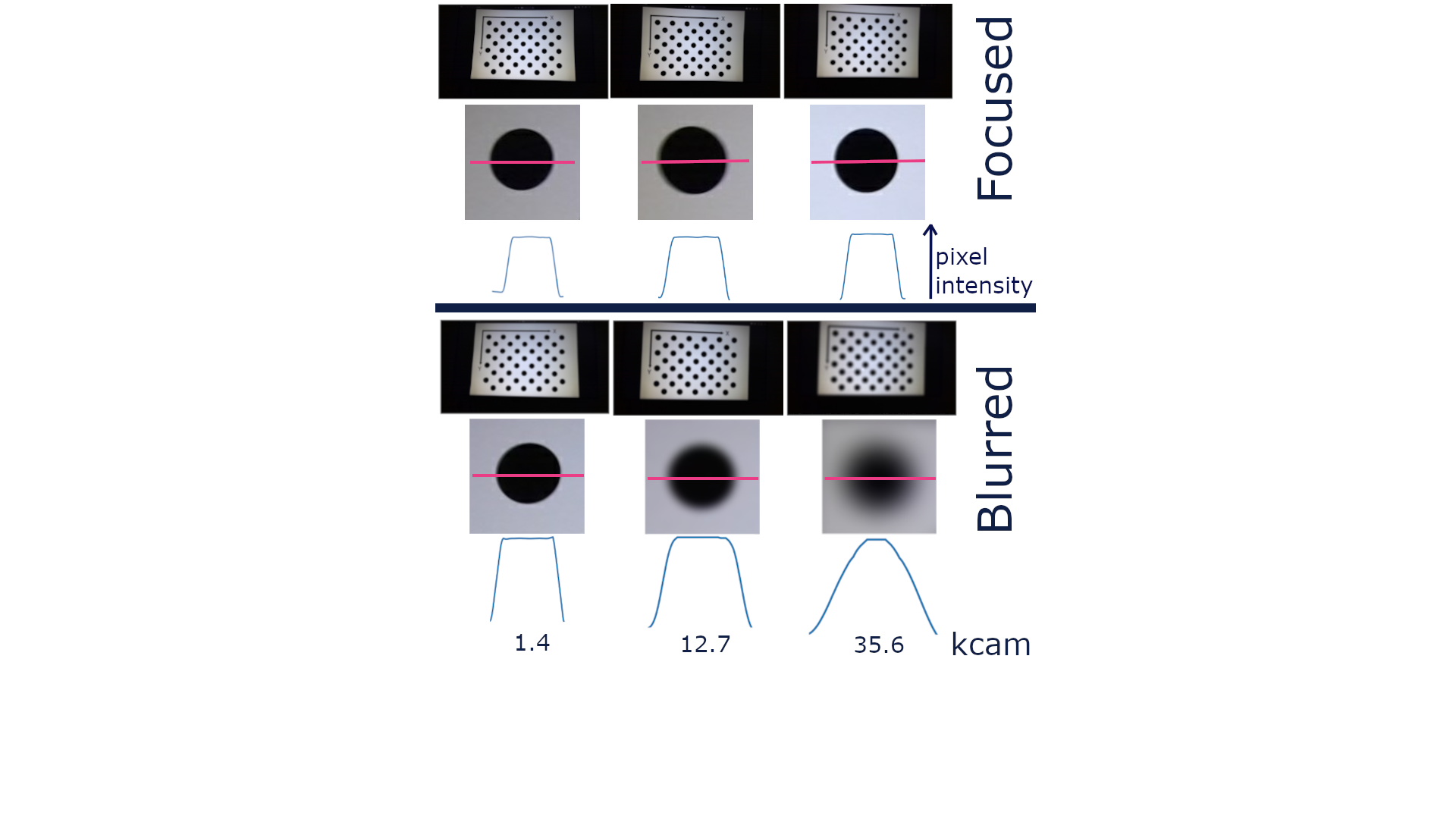}
   \vspace{-2cm}
   \caption{Defocus Blur Calibration}
      \vspace{-0.5cm}
\end{figure}

\section{Experiments}
\subsection{Datasets}

\noindent \textbf{Defocusnet dataset.}
We use the synthetic dataset generated by Maximov et al. \cite{Maximov2020FocusEstimation} to train one of our models. This dataset was created with a virtual camera having several $K_{cam}$ values of 0.15,0.33,0.78,1.59 and 2.41. This dataset has 500 focal stacks, each with 5 images with different focal distances. 

\noindent \textbf{Synthetic Blender dataset.}
We create a new synthetic dataset by expanding the defocusnet dataset \cite{Maximov2020FocusEstimation}. This new dataset has various textures (to make them realistic) mapped to the 3D objects that was not present in the original dataset. We use several simulated cameras with $K_{cam}$ of 0.08, 0.15, 0.23 and 0.33. This dataset has a focal distance of 1.5m and contains 400 defocus blurred images. We use the script provided by Maximov et al. (modified) to create our dataset. The images we generate in this dataset are 256 x 256 pixels.

Both the Synthetic blender and the defocusnet datasets also have a perfectly focused image per each defocus blurred image. 

\noindent \textbf{DDFF12 dataset.} We also use the DDFF12 dataset provided by Hazirbas et al. \cite{Hazirbas2019DeepFocus} which contains 720 images created with a lightfield camera. We use the two real world datasets (DDFF12 and the NYU dataset described next) so that we can show that our models can work under real world images and deal with the domain gap between real and synthetic images \cite{Xian2020Structure-guidedPrediction}. After pre-processing the images as mentioned by Hazirbas et al. we obtained the blurred images which are focused at various distances.

\noindent \textbf{NYU depth v2 dataset.}
The NYU depth v2 dataset \cite{Silberman2012IndoorImages.} contains 1449 pairs of aligned RGB and depth image pairs. Following previous papers \cite{Maximov2020FocusEstimation} \cite{Eigen2014DepthNetwork} we create the training and testing splits. We create artificially defocus blurred images from this dataset using the method described by Carvalho et al. \cite{Carvalho2018DeepNetworks}. We have fixed certain drawbacks in their Matlab script in order to produce more realistic defocus blurred images as further discussed in the appendix. We used $K_{cam}$ values for training and testing as shown in Table 3. Images of 480 x 480 pixels were used for training and 480 x 640 were used for testing. 

\subsection{Experimental Setup}
We use PyTorch \cite{Paszke2019Pytorch:Library} to implement the neural networks. We use the Adam optimizer \cite{Kingma2014Adam:Optimization} with $\beta_1=0.9$ and $\beta_2=0.999$ and a learning rate of $10^{-4}$. Mean Squared Error was used as the loss function for both the blur and the depth to train all the models. We evaluate our depth predictions with the metrics absolute relative error (REL), mean-squared error (MSE), Root-Mean-Squared error (RMSE) and average log10 error. We also report threshold accuracy $\delta_n$ which is the percentage of pixels which satisfy the condition $max(d_i/\hat{d_i},\hat{d_i}/d_i)<1.25^n$. We train our models on the defocusnet dataset for 400 epochs and a batch size of 20. Our NUY depth models were train for 800 epochs with a batch size of 8.

\subsection{Performance}

\begin{table}
  \centering
\begin{tabular}{l|c c c c}
            \hline
            Method & \multicolumn{4}{c@{}}{$K_{cam}$}\\
            &0.08 & 0.14 & 0.23 & 0.33 \\ 
            \hline
            in-focus &0.099&0.081&0.082&0.100\\
            No $K_{cam}$ &0.062&0.050&0.056&0.085\\
            GT $K_{cam}$  &0.045&0.037&0.052&0.061\\
            \hline
    \end{tabular}
  \caption{Performance on Blender dataset (MSE)}
     \vspace{-0.5cm}
\end{table}

Table 2 shows the performance of the model trained on the defocusnet \cite{Maximov2020FocusEstimation} dataset and evaluated on our Blender dataset with different $k_{cam}$ values of simulated cameras. All three methods; in-focus, No $K_{cam}$ and GT $K_{cam}$, use the same deep learning architecture to predict depth. The only exception is that the GT $K_{cam}$ model performs the $K_{cam}$ correction as shown in Figure 2. We use the $K_{cam}$ values that were used to generate the data and these can be called the Ground Truth $K_{cam}$ values (GT $K_{cam}$). The In-focus model was both trained and tested on perfectly focused images. The No $K_{cam}$ model does not consider the effect of $K_{cam}$ during either training or testing (similar to defocusnet \cite{Maximov2020FocusEstimation} model). This means the No $K_{cam}$ model does not divide the output of the blur estimation model with $K_{cam}$ as shown in Figure 2. The GT $K_{cam}$ model on the other hand considers the effect of $K_{cam}$ and behaves as shown in Figure 2 during both training and testing. According to Table 2 the performance of both the No $K_{cam}$ and the GT $K_{cam}$ models are better (by around 0.025) than that of the in-focus method which shows that considering defocus blur is valuable when estimating depth. Our models perform better when considering the effect of $K_{cam}$ (GT $K_{cam}$) compared to when not considering it (No $K_{cam}$) by around 0.015 in MSE. This shows that we can transfer the knowledge learned with the trained model into a new domain (images taken with a different $K_{cam}$) just with one parameter $K_{cam}$.

\begin{table}
  \centering
  \resizebox{0.5\textwidth}{!}{%
  \begin{tabular}{c c c c c c c c}
    \toprule
    $K_{cam}$ & method & $\delta_1 \uparrow$  & $\delta_2 \uparrow$& $\delta_3 \uparrow$ & REL $\downarrow$ & RMSE $\downarrow$ & log10 $\downarrow$\\
    \midrule
    in-focus & VPD\cite{Zhao2023UnleashingPerception} & 0.953 & 0.992 & 0.999 & 0.052 & 0.154 & 0.027\\
    \midrule
    
    8.79 & GT $K_{cam}$ & 0.976 &  0.997 & 0.999 & 0.046 & 0.082 & 0.019 \\
    8.79 & No $K_{cam}$ & 0.912 & 0.975 &  0.998 & 0.095 & 0.161 & 0.037 \\
    35.61 & GT $K_{cam}$ & 0.976 & 0.997 & 0.999 & 0.046 & 0.082 & 0.019\\
    35.61 & No $K_{cam}$ & 0.962 & 0.995 & 0.999 & 0.054 & 0.101 & 0.023 \\
    \midrule
    
    12.69 & GT $K_{cam}$ & 0.969 & 0.999 & 0.999 & 0.068 & 0.123 & 0.088 \\
    12.69 & est $K_{cam}$ &  0.970 & 0.999 & 0.999 & 0.068 & 0.122 & 0.030 \\
    12.69 & No $K_{cam}$ & 0.853 & 0.963 & 0.999 & 0.127 & 0.193 & 0.050 \\

    22.67 & GT $K_{cam}$ & 0.980 & 0.998 & 0.999 & 0.068 & 0.117 & 0.028 \\
    22.67 & est $K_{cam}$ & 0.980 & 0.998 & 0.999 & 0.069 & 0.118 & 0.028 \\
    22.67 & No $K_{cam}$ & 0.896 & 0.994 & 0.999 & 0.105 &  0.165 & 0.043 \\
    
    \bottomrule
  \end{tabular}}
  \caption{Performance on NYU dataset}
 \vspace{-0.5cm}
\end{table}

Table 3 shows the performance on the defocus blurred NYU depth dataset. Here we use a single trained model to evaluate the performance under various settings. The model was trained on data refocused with a $K_{cam}$ of 8.79 and 35.61 and tested on the rest under the distance range of 0 to 2 m. The VPD model was trained and tested on in-focus images with no defocus blurring. Our GT and est $K_{cam}$ methods outperform the state-of-the-art depth estimation model (VPD) on the NUY depth v2 dataset \cite{Zhao2023UnleashingPerception} by around 0.04 in RMSE. This converts to a reduction of error of around 4cm in the depth estimation. This proves again the importance of defocus blurring in depth estimation. We evaluate our models under three methods which depend on the nature of the $K_{cam}$ values used. The method column, "GT $K_{cam}$" means we have used the $K_{cam}$ values that were used to defocus blur the particular dataset which can be considered as Ground Truth $K_{cam}$ values. "est $K_{cam}$" represents the $K_{cam}$ values that were estimated with the defocus calibration method described in section 3.3. No $K_{cam}$ models do not consider the effect of $K_{cam}$. We describe further details of the estimation process in the subsequent sections.

Table 4 shows the performance of our model under the DDFF12 dataset \cite{Hazirbas2019DeepFocus}. All the models were first trained on the defocusnet dataset \cite{Maximov2020FocusEstimation} (the same model we used to evaluate the Blender dataset as shown in Table 2). The In-focus model was trained and tested on well focused images. The No $K_{cam}$ and est $K_{cam}$ models were trained and tested on defocus blurred images. Since we do not have ground truth $K_{cam}$ for the DDFF12 dataset, we performed a linear search of the $K_{cam}$ which predicts the best depth using the ground truth depth maps provided in the training set. The results in Table 4 are the performance of our model under the test set using the $K_{cam}$ value found above. Both the No $K_{cam}$ and est $K_{cam}$ models perform better than the in-focus model for depth prediction. Also using the appropriate $K_{cam}$ to transfer the model to the new domain of images significantly improves the performance compared to the no $K_{cam}$ model which does not perform a correction that depends on the camera. 

\subsection{Defocus Blur Calibration Performance}

\begin{figure}
   \includegraphics[width=\columnwidth]{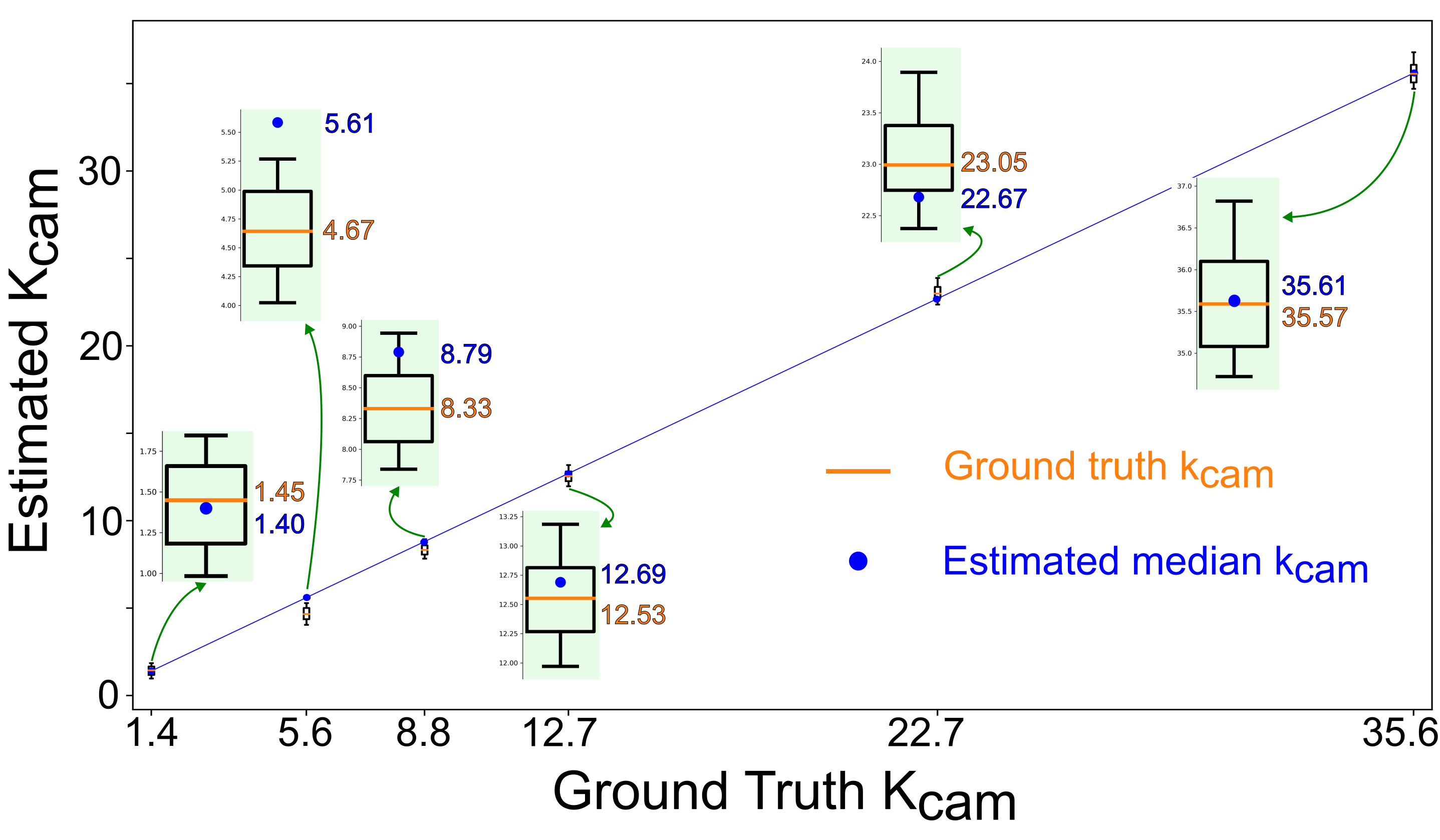}
   \caption{Estimation of $K_{cam}$ values of different cameras}
      \vspace{-0.5cm}
\end{figure}

We expand the discussion on defocus blur calibration in this section. These experiments were performed on the refocused NYU depth v2 dataset. We have created refocused data with $K_{cam}$ values of 1.39, 5.61, 8.79, 12.69, 22.67, 25.61. Note that we have used the additional $K_{cam}$ values (1.39 and 5.61) that were not used to evaluate the performance of depth estimation in Table 2. We obtain several photos of the asymmetric circular pattern shown in Figure 3 with the Microsoft Kinect camera and refocus them with the above mentioned $K_{cam}$ values. We used from 19 to 20 different image pairs (an in-focus image and a defocus-blurred image) for each $K_{cam}$ value. Then we perform the defocus blur calibration procedure described in section 3.2. Estimated $K_{cam}$ values vs. the actual values (ground truth $K_{cam}$) are shown in Figure 4. The relationship between the ground truth and estimated $K_{cam}$ values are very linear as expected. We estimate one $K_{cam}$ value per one circle from an in-focus and defocus blurred image pair. Since there are 44 circles in the pattern, for 20 image pairs we obtained 880 estimated $K_{cam}$ values. The results in Figure 4 were obtained after removing outliers and calculating the median from the estimated $K_{cam}$ values. We show box plots with interquartile range, median, minimum and maximum values of these estimations along with the ground truth $K_{cam}$ values. 
\vspace{-0.6cm}
\subsubsection{Sensitivity of depth estimation performance to $K_{cam}$}
In this section we explore how the variation in estimated $K_{cam}$ values affect the depth estimation performance. We use the same model that we used to obtain the results in Table 3 that was trained on data from $K_{cam}$ values of 8.79 and 35.61 and evaluate them on data from $K_{cam}$ values of 12.69 and 22.67. As can be seen in Figure 5, we use a range of numbers centered on the actual $K_{cam}$ values for the respective datasets and obtain the RMSE error of depth estimation. It can be seen that the error response of the model to the variation of $K_{cam}$ used has a clear minimum. The error increases if the values used in the place of $K_{cam}$ deviates from the actual value. For example, the error of the response of $K_{cam}$=23.67 increases by around 16\% if the $K_{cam}$ used deviates positively from the GT values by 18\%. Figure 8 shows some examples of predicted depth maps when the model was provided with an unseen virtually blurred image from a camera with $K_{cam}=22.67$. Agreeing with the Figure 5, the predictions get distorted faster when the $K_{cam}$ used lowers than the ground truth $K_{cam}$ and distorts slower when it increases.

\begin{figure}
\centering
   \includegraphics[width=0.7\columnwidth]{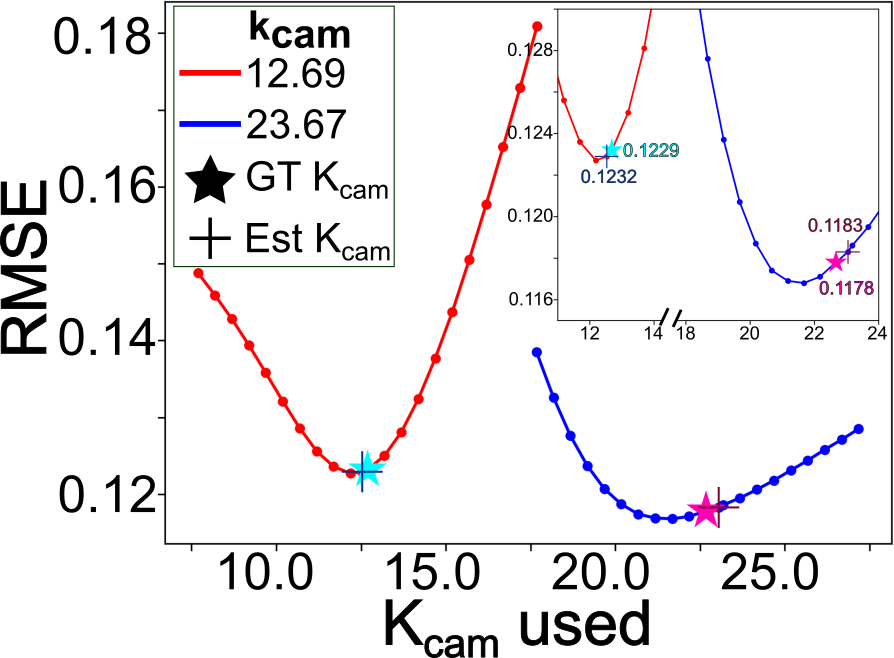}
   \caption{$K_{cam}$ Estimation Error}
\end{figure}

\subsection{Effect of the blur weight}
We change the scaling parameter $b\textunderscore weight$ from equation 8 while training several models on data from defocus blurred NYU depth v2 dataset with $K_{cam}$ values of 8.79 and 35.61. The performance on the two evaluation datasets (with $K_{cam}$ values of 12.69 and 22.67) are shown in Figure 7. 

\subsection{Effect of the Field of View}

\begin{figure}
  \begin{minipage}{.4\columnwidth}
  \centering
    \includegraphics[width=0.6\columnwidth]{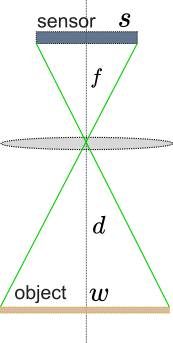}
   \caption{FOV of a camera}
   \vspace{-0.5cm}
  \end{minipage}
  \hfill
  \begin{minipage}{.6\columnwidth}
  \centering
    \includegraphics[width=1\columnwidth]{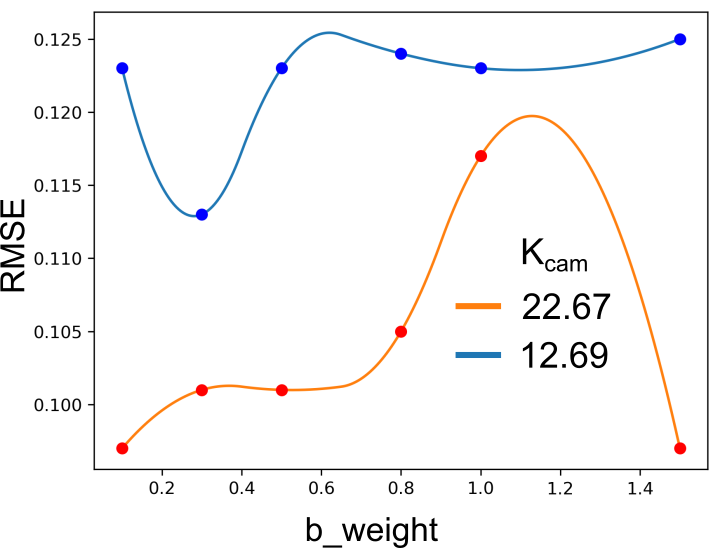}
   \caption{RMSE Vs b\textunderscore weight}
   \vspace{-0.6cm}
  \end{minipage}
\end{figure}

\begin{table}
\vspace{0.3cm}
    \parbox{.40\linewidth}{
        \centering
        \resizebox{1\linewidth}{!}{%
        \begin{tabular}{l|c}
        \hline
        Method &MSE \\
        \hline
        in-focus & 0.0640 \\
        No $K_{cam}$ &  0.0139 \\
        est $K_{cam}$ & 0.0096 \\
        \hline
        \end{tabular}}
        \caption{Performance on DDFF12 dataset}
        \vspace{-0.3cm}
    }
    \hfill
    \parbox{.58\linewidth}{
        \centering
        \resizebox{0.9\linewidth}{!}{%
        \begin{tabular}{l|c c}
        \hline
        $K_{cam}$ & Image size & RMSE \\
        \hline
        12.69 & Original size & 0.123 \\
              & Resized & 0.438 \\
        \hline
        22.67 & Original size & 0.117 \\
              & Resized & 0.399\\
        \hline
\end{tabular}}
  \caption{Effect of FOV}}
  \vspace{-0.3cm}
\end{table}

Field of view of a camera can be defined in several ways. One way is to define it as the size of an object at a given distance from the camera that would completely fill the image sensor. In Figure 6, $s$ is the length of the sensor, f is the focal length of the lens, $d$ is the distance to the object and $w$ is the length of the object. The size $w$ of an object that would completely fill the sensor will be given by $w = \frac{s}{f} \cdot d$. It can be seen that $w$ is inversely proportional to $f$. Cameras with a smaller focal length have a larger Field of View and vise-versa. In all the experiments that we performed including the NYU dataset, we have assumed that the cameras have a fixed FOV even when the $k_{cam}$ (and therefore $f$) changes. While this is helpful to analyze the performance of blur based depth estimation methods, it is important to investigate the effect of the FOV change on the performance of the models. We created a dataset by scaling down the images in the NYU depth dataset by a factor of 0.6 and then refocusing with a $k_{cam}$ (respective $f$ is 30mm) of 12.69. The model has been trained with data having $k_{cam}$s of 8.79 and 35.61 (they had focal lengths ($f$) of 20mm and 50mm). We scaled down the images of $f=30mm$ with respect to $f=50mm$ which is 0.6. Note that the images have the same amount of blur as the images of original size; only the size of the objects visible have changed. From Table 5, it can be seen that the performance drops significantly by more than three folds when we perform the resizing. The reason for this can be understood from Figure 2. The depth estimation section receives two inputs. One is the blur and the second is the image features in the form of skip connections. Although we account for the change of blur through division by the respective $k_{cam}$, we do not modify image features to reflect the change of FOV. This is a limitation of our work and needs to be addressed in the future. 

\section{Conclusions}

\begin{figure}
\centering
   \includegraphics[width=0.9\columnwidth]{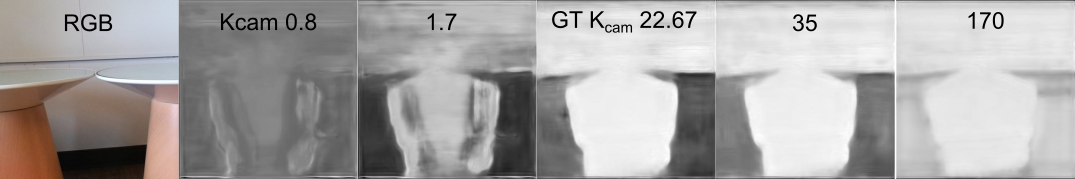}
   \caption{Examples of from the camera with $K_{cam}=22.67$ predicted using various $K_{cam}$ values}
   \vspace{-0.5cm}
\end{figure}
We show that estimating depth from defocus blur is significantly superior to conventional semantic based depth prediction provided that the camera is suitable for it. But this technique is sensitive to the camera. Our novel approach performs a simple correction to an already trained depth prediction model using camera parameters of a given camera. We show that this correction can alleviate the sensitivity of the model to the camera. Our novel defocus blur calibration technique can estimate the camera parameters using several images taken by a given camera. We show that our approach beats the state-of-the art for several datasets. Finally we show some limitations of our work and suggest future improvements.  

Methods such as ours which utilize a camera can come with certain privacy and security concerns. Our technique specifically can be used by enemy drones to measure distance to a target person in close quarters and cause significant harm which raises several ethical concerns. As researchers it is important to address these concerns alongside the algorithmic improvements to the state-of-the-art. 

\textbf{Acknowledgment:} This work is supported by the award 70NANB21H029 from the U.S. Department of Commerce, National Institute of Standards and Technology (NIST).

\FloatBarrier

\bibliographystyle{IEEEtran}
\bibliography{main}

\end{document}